\documentclass[10pt,twocolumn,letterpaper]{article}

\usepackage[pagenumbers]{cvpr} 

\usepackage{graphicx}
\usepackage{amsmath}
\usepackage{amssymb}
\usepackage{booktabs}
\usepackage{adjustbox}

\graphicspath{ {images/} }

\usepackage[pagebackref,breaklinks,colorlinks]{hyperref}

\usepackage[capitalize]{cleveref}
\crefname{section}{Sec.}{Secs.}
\Crefname{section}{Section}{Sections}
\Crefname{table}{Table}{Tables}
\crefname{table}{Tab.}{Tabs.}

\def\confName{NLP}
\def\confYear{2024}

\begin{document}

\title{Sentiment Analysis Across Languages: Evaluation Before and After Machine Translation to English}

\author{Aekansh Kathunia\\
IIIT Delhi\\
{\tt\small aekansh21127@iiitd.ac.in}
\and
Mohammad Kaif\\
IIIT Delhi\\
{\tt\small kaif21067@iiitd.ac.in}
\and
Nalin Arora\\
IIIT Delhi\\
{\tt\small nalin21478@iiitd.ac.in}
\and
N Narotam\\
IIIT Delhi\\
{\tt\small narotam21477@iiitd.ac.in}
}
\maketitle

\begin{abstract}
People communicate in more than 7,000 languages around the world, with around 780 languages spoken in India alone. Despite this linguistic diversity, research on Sentiment Analysis has predominantly focused on English text data, resulting in a disproportionate availability of sentiment resources for English. This paper examines the performance of transformer models in Sentiment Analysis tasks across multilingual datasets and text that has undergone machine translation. By comparing the effectiveness of these models in different linguistic contexts, we gain insights into their performance variations and potential implications for sentiment analysis across diverse languages. We also discuss the shortcomings and potential for future work towards the end.
\end{abstract}

\section{Introduction}
\label{sec:intro}

The term \textit{Sentiment Analysis} refers to the process of analyzing text to determine the emotional tone of the message. More generally, it can be understood as assessing an individual towards a particular target.

\textit{Machine translation} refers to the conversion of text from a source language to a target language via the use of computer algorithms.

\vspace{3mm}
Bert and XLM Roberta are Large language models based on the transformer architecture introduced by Google, trained on a huge corpus of data, ideal for fine-tuning downstream tasks such as Sentiment Analysis and Machine Translation. 
Given the accessibility and usefulness of these models, they find their application in various languages other than English and various multilingual settings, where models are tuned to understand context in multiple languages.

\vspace{5mm}

Due to easy accessibility and the plethora of literature available in English, tasks like \textit{Sentiment Analysis} have extensively been researched on English texts, which naturally leads to many sentiment resources for English texts but less so for texts in other languages.

\vspace{5mm}

In our project, we aim to compare \textit{Sentiment Analysis} performance on the original language texts for the languages French, German, Spanish, Japanese and Chinese while also comparing machine translation performance of models across different languages. 

\vspace{5mm}

Other work done with similar objectives, we believe, often falls short of creating a robust pipeline to process multilingual datasets and often implements simple rudimentary pipelines that don't use underlying datasets to the fullest. We identify certain gaps and interesting areas in which we could expand existing research done on this topic and thus present the major contributions of our project:
\begin{itemize}
    \item We present robust pipelines that incorporate and compare various state-of-the-art sentiment analysis and machine translation models.
    \item We provide domain-tuned versions of large language models on a subset of the \textbf{Multilingual Amazon Reviews Corpus}\cite{keung_multilingual_amazon}.
    \item We analyse the translation models in different languages by their ability to recreate the baseline for sentiment analysis of English models. This allows us to understand the progress of NLP in different languages compared to English.
    \item We explore if it is viable to use cross-lingual over uni-lingual models and if significant performance can be achieved by machine translation to transform the dataset into English, in which models are, in general, better.
\end{itemize}
For all tasks, we use transformer-based models that have been pre-trained on an enormous corpus of text prior to our deployment and pertaining. 

\section{Related Work}

\begin{figure*}
    \centering
    \includegraphics[width=\textwidth]{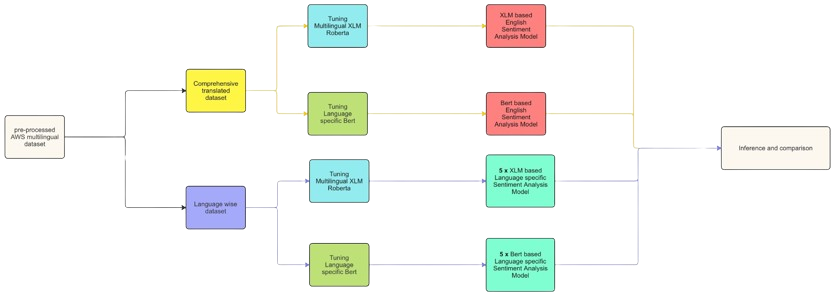}
    \caption{Training Pipeline}
    \label{fig:your_label}
\end{figure*}

\subsection{Sentiment Analysis and Emotion Detection from Text}
There have been several studies exploring sentiment analysis and emotion detection techniques being applied to textual data. The paper from Nandwani and Verma provides insights into different methodologies used for analyzing sentiments and detecting emotions. They delve into traditional machine learning algorithms and deep learning models, highlighting their strengths and limitations on the above-mentioned task.\cite{nandwani_verma}.

\vspace{5mm}

Mohammad et al. examine how translation alters sentiment, focusing on the impact of machine translation on sentiment analysis across languages \cite{mohammad_translation}. They observe that machine translation can significantly alter the sentiment expressed in a sentence and highlight the challenges of accurately capturing sentiment in a multilingual setting.

\vspace{5mm}

Araujo et al. evaluate machine translation for multilingual sentence-level sentiment analysis and assess the effectiveness of different machine translation models in preserving sentiment across languages \cite{araujo_translation}. However, their findings underscore the importance of considering the quality of machine translation while conducting sentiment analysis in multilingual contexts.

\subsection{Challenges in Multilingual Sentiment Analysis}
One of the key challenges identified by both Mohammad et al. and Araujo et al. in accurately interpreting emotions from text is the inherent complexity of language, which includes nuances, context, and cultural disparities. This challenge is more inherent in a multilingual setting, where the translation of text to English may not fully capture the original sentiment or emotion.

\vspace{5mm}

Moreover, recent advancements in sentiment analysis and emotion detection have focused on integrating multimodal data and leveraging pre-trained language models to enhance accuracy. These developments are particularly relevant in the context of multilingual sentiment analysis, where incorporating additional modalities such as images or audio can provide valuable context for understanding emotions expressed in text.

\vspace{5mm}

Overall, the insights provided by Nandwani and Verma, Mohammad et al., and Araujo et al. offer valuable considerations for evaluating sentiment analysis across languages, both before and after machine translation to English.

\subsection{BERT}
BERT (Bidirectional Encoder Representations from Transformers) is a pre-trained language model developed by Google \cite{devlin-etal-2019-bert}. It is based on a multi-layer bidirectional transformer encoder, which generates contextualized representations of input text. BERT is pre-trained on a large corpus of text and is fine-tuned on specific tasks, achieving state-of-the-art results in various natural language processing tasks.

For languages other than English, we used the following instances of Bert from hugging face:

1. bert-base-german-cased\cite{chan-etal-2020-germans} - This is a German language model based on the BERT architecture. It was developed by Google and has been fine-tuned on a large corpus of German text.

2. dccuchile/bert-base-spanish-wwm-cased\cite{CaneteCFP2020} - This is a Spanish language model based on the BERT architecture. It was developed by the University of Chile and has been fine-tuned on a large corpus of Spanish text.

3. dbmdz/bert-base-french-europeana-cased\cite{french-bert} - is a French language model based on the BERT architecture. It was developed by the researchers at the Center for Information and Language Processing (CIS), LMU Munich.

4. cl-tohoku/bert-base-japanese\cite{japanese-bert} - This is a Japanese language model based on the BERT architecture. It was developed by Tohoku University and has been fine-tuned on a large corpus of Japanese text.

5. bert-base-chinese(addition of original bert paper \cite{chinese_bert} - This is a Chinese language model based on the BERT architecture. It was developed by Google and has been fine-tuned on a large corpus of Chinese text.

\vspace{5mm}

BERT has been shown to be effective in sentiment classification tasks, achieving high accuracy by leveraging its contextualized representations and fine-tuning on specific datasets \cite{Sayeed2023-ir}. BERT fine-tuning has led to remarkable state-of-the-art results on various downstream tasks, including sentiment analysis.

\subsection{XLM RoBERTa}
XLM-RoBERTa\cite{Conneau2019-th} is a multi-lingual language model that combines the strengths of XLM \cite{NEURIPS2019_c04c19c2} and RoBERTa \cite{Liu2019-pz}. The architecture is based on the RoBERTa model, with a modified XLM encoder that enables cross-lingual transfer learning. This allows XLM-RoBERTa to leverage pre-training in multiple languages and fine-tune on specific tasks, achieving state-of-the-art results in various natural language processing tasks.

\vspace{5mm}

XLM-RoBERTa has been shown to be effective in sentiment classification tasks, achieving high accuracy in multiple languages \cite{barbieri-etal-2022-xlm}. By leveraging its multi-lingual capabilities and fine-tuning on our dataset, we achieve high performance in sentiment classification.

\section{Methodology}

We meticulously constructed a robust pipeline to fulfill the project's objectives of comparing transformer performance in sentiment analysis across multilingual data and machine-translated text. This pipeline used advanced transformer architectures such as BERT and XLM-RoBERTa.

\vspace{5mm}

BERT, short for Bidirectional Encoder Representations from Transformers, is an encoder-only model that utilises masked language modeling and next-sentence prediction techniques. BERT models are primarily trained in a single language context.

\vspace{5mm}

XLM-RoBERTa is an advancement over XLM, leveraging the robust RoBERTa architecture. Through extensive pre-training on a larger dataset and prolonged training sessions, XLM-RoBERTa excels in various NLP tasks. It inherits XLM's cross-lingual capabilities and benefits from RoBERTa's enhanced representation learning, making it highly proficient in handling multilingual data and achieving superior performance across diverse NLP tasks.

\vspace{5mm}

To achieve our machine translation objective, we employed the OPUS-MT machine translation model, which utilizes state-of-the-art transformer-based neural machine translation techniques. By leveraging the usage of transformers, OPUS-MT achieves good-quality translations and fluency across multiple languages. This choice aligns perfectly with our project's focus on evaluating transformer performance on machine-translated text, ensuring robustness and reliability in our analyses.
\vspace{5mm}

To compare the model performance, we utilized F1-scores as the metric since they give a holistic review of the model's performance across classes.

\vspace{5mm}

The pipeline we used during the project is as follows, we first used 50000 entries from each language due to limited computational capability to fine-tune BERT and XLM-RoBERTa models and evaluate their performance using F1-score. In the second part we translated 20000 entries from each language using OPUS-MT to English and then combined them together to form a dataset of 100000 entries. We fine-tuned our BERT and XLM-RoBERTa models on this combined dataset and then evaluated the model's performance using the F1-score as the evaluation metric.

\vspace{5mm}

By employing these state-of-the-art transformer models within our pipeline, we aimed to comprehensively evaluate their efficacy in sentiment analysis tasks across multilingual datasets and machine-translated texts. 

\begin{table*}
\centering
\hsize=\textwidth
\caption{Training corpus statistics. 200,000 reviews per language. Taken from the corpus review paper by Keung et al.}
\begin{tabular}{lrrrrrr}
\toprule
 & En & De & Es & Fr & Ja & Zh\\
\midrule
Number of products & 196,745 & 189,148 & 179,076 & 183,345 & 185,436 & 164,540\\
Number of reviewers & 185,541 & 171,620 & 150,938 & 157,922 & 164,776 & 132,246\\
Average characters/review & 178.8 & 207.9 & 151.3 & 159.4 & 101.4 & 51.0\\
Average characters/review title & 24.2 & 21.8 & 19.2 & 19.1 & 9.5 & 7.6\\
\bottomrule
\end{tabular}
\end{table*}

\section{Dataset}




The dataset used in this study is the \textbf{Multilingual Amazon Reviews Corpus}, as described by Keung et al. \cite{keung_multilingual_amazon}. This corpus represents a rich collection of product reviews gathered from the Amazon platform, spanning multiple languages, including English, Spanish, French, German, Japanese, Chinese, and Italian.

\vspace{5mm}

The Multilingual Amazon Reviews Corpus contains reviews across a wide range of product categories, such as electronics, books, movies, home appliances, and more. Each review entry within the dataset is accompanied by comprehensive metadata, including the product ID, reviewer ID, review text, star rating, and review date. Additionally, the dataset contains information about the geographic location of reviewers, providing insights into regional variations in sentiment expression.

\vspace{5mm}

One of the notable features of the Multilingual Amazon Reviews Corpus is its extensive coverage of languages and product categories, making it a valuable resource for studying sentiment analysis and multilingual natural language processing tasks. Researchers can leverage this dataset to explore the complexities and nuances of sentiment analysis across diverse linguistic and cultural contexts.

\vspace{5mm}

In this study, we selected the Multilingual Amazon Reviews Corpus due to its multilingual nature and diverse product categories, which provided us with an opportunity to investigate the challenges and opportunities of sentiment analysis across different languages and domains. 

\section{Experimental Setup}

Due to limited computational resources, we utilize a randomly sampled subset of the Multilingual Amazon Reviews Corpus, consisting of 50,000 samples of the languages French, German, Spanish, Japanese and Chinese while also ensuring that the data is balanced.

\begin{table}[h]
\centering
\label{tab:model_mapping}
\begin{adjustbox}{width=0.45\textwidth}
\begin{tabular}{lc}
\hline
Hugging Face Model ID & Language Specific \\
\hline
bert-base-german-cased & german\_BERT \\
dccuchile/bert-base-spanish-wwm-cased  & spanish\_BERT \\
dbmdz/bert-base-french-europeana-cased & french\_BERT \\
cl-tohoku/bert-base-japanese & japanese\_BERT \\
bert-base-chinese & chinese\_BERT \\
bert-base-uncased  & english\_BERT \\
xlm-roberta-base  & Multi-Lingual \\
\hline
\end{tabular}
\end{adjustbox}
\caption{Mapping Hugging Face Model IDs to Language Specificity}
\end{table}

We tune uni-lingual BERT models for each language and also a multi-lingual XLM-RoBERTa model on our dataset to further assess the performance of these different languages across different languages. The performance of each model in each language is ultimately compared to the performance of BERT and XLM-RoBERTa on English models tuned on the dataset.

\vspace{5mm}

For our evaluation, we observe the model performance on two tasks:
\begin{enumerate}
    \item \textbf{Sentiment Classification:} Classify the review text into two classes, positive review and negative review.
    \item \textbf{Star Rating Prediction:} Predict the star rating of the review text.
\end{enumerate}

\begin{figure}[h]
\includegraphics[width=3.5cm, height=4cm]{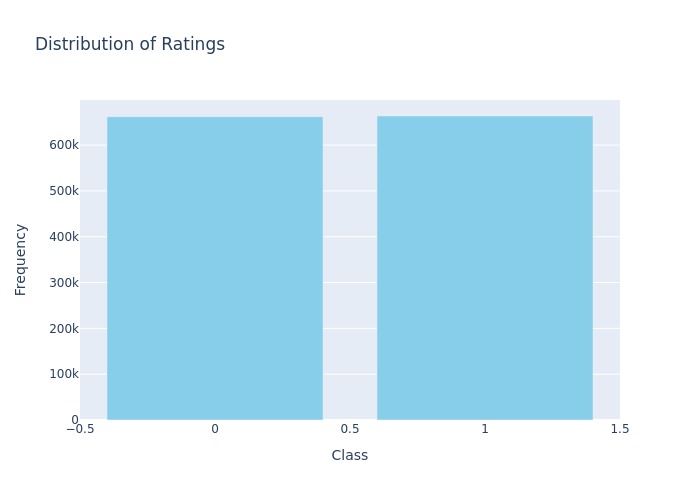}
\hspace{1cm}
\includegraphics[width=3.5cm, height=4cm]{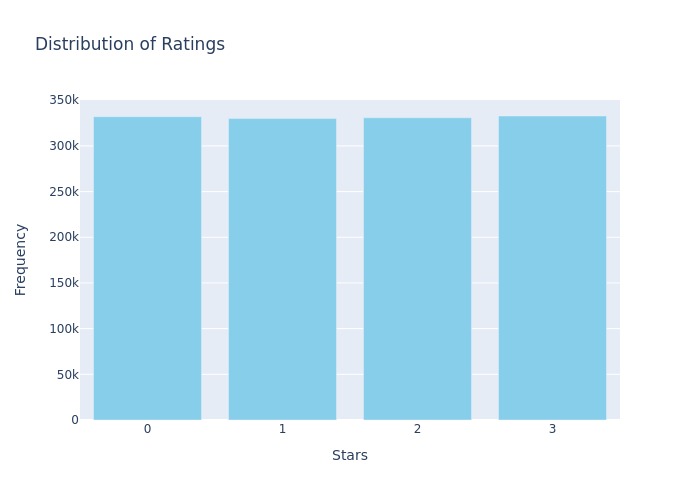}
\caption{Distribution of Labels}
\end{figure}

The gold labels for Star Rating Prediction were already present in the dataset. However, the gold labels for Sentiment Classification were not present in the dataset and were obtained from the annotated "hungnm/multilingual-amazon-review-sentiment-processed" \cite{keung_multilingual_amazon_huggingface} dataset from huggingface.

\vspace{5mm}

The model evaluation tasks were first performed on all of the sampled data. Then, 20,000 samples from each of the source languages were translated to the target language, English. The evaluation tasks were then repeated on the translated samples to assess post \textit{Machine Translation} performance.


\section{Results}
\label{sec:formatting}

\begin{table}[h]
	
        \footnotesize
        \centering
    \begin{adjustbox}{width=0.45\textwidth}
    \begin{tabular}{|c|c|c|} 
        \hline
        Language & Before Machine Translation & After Machine Translation \\
        \hline
        \textbf{Spanish(ES)} & 
        \begin{tabular}{cc}
       \textbf{{Spanish\_XLM}} & 0.90461 \\
       \textbf{{Spanish\_BERT}} & 0.88878 \\
    \end{tabular} &
    \begin{tabular}{cc}
        \textbf{{Spanish\_XLM}} &  0.89965 \\ 
        \textbf{{Spanish\_BERT}} & 0.89875 \\
    \end{tabular} \\ \hline
    
         \textbf{German (DE)} & 
        \begin{tabular}{cc}
       \textbf{{German\_XLM}} &  0.90585 \\
       \textbf{{German\_BERT}} & 0.90730 \\
    \end{tabular} &
    \begin{tabular}{cc}
        \textbf{{German\_XLM}} &  0.89420 \\ 
        \textbf{{German\_BERT}} & 0.89672 \\
    \end{tabular} \\ \hline
    
     \textbf{French (FR)} & 
        \begin{tabular}{cc}
       \textbf{{French\_XLM}} &  0.90294 \\
       \textbf{{French\_BERT}} & 0.87775 \\
    \end{tabular} &
    \begin{tabular}{cc}
        \textbf{{French\_XLM}} &  0.88740 \\
        \textbf{{French\_BERT}} & 0.89131 \\
    \end{tabular} \\ \hline
    
     \textbf{Chinese (ZH)} & 
        \begin{tabular}{cc}
       \textbf{{Chinese\_XLM}} &  0.86715 \\
       \textbf{{Chinese\_BERT}} & 0.86844 \\
    \end{tabular} &
    \begin{tabular}{cc}
        \textbf{{Chinese\_XLM }} &  0.83660 \\      
        \textbf{{Chinese\_BERT}} & 0.82391 \\
    \end{tabular} \\ \hline
    
     \textbf{Japanese(JA)} & 
        \begin{tabular}{cc}
       \textbf{{Japanese\_XLM}} &  0.90161 \\
       \textbf{{Japanese\_BERT}} & 0.89231 \\
    \end{tabular} &
    \begin{tabular}{cc}
        \textbf{{Japanese\_XLM}} &  0.83049 \\     
        \textbf{{Japanese\_BERT}} & 0.82479\\
    \end{tabular} \\ \hline

        \hline
    \end{tabular}
    \end{adjustbox}
    \caption{Sentiment Classification}
    \label{tab:SAt}
\label{table:SA}
\end{table}

\begin{table}[h]
	
        \footnotesize
        \centering
    \begin{adjustbox}{width=0.45\textwidth}
    \begin{tabular}{|c|c|c|} 
        \hline
        Language & Before Machine Translation & After Machine Translation \\
        \hline
        \textbf{Spanish(ES)} & 
        \begin{tabular}{cc}
        
       \textbf{{Spanish\_XLM}} & 0.60697 \\
       \textbf{{Spanish\_BERT}} & 0.58337\\
    \end{tabular} &
    \begin{tabular}{cc}
        \textbf{{Spanish\_XLM}} &  0.58155 \\ 
        \textbf{{Spanish\_BERT}} & 0.62267 \\
    \end{tabular} \\ \hline
    
         \textbf{German (DE)} & 
        \begin{tabular}{cc}
       \textbf{{German\_XLM}} &  0.64942 \\
       \textbf{{German\_BERT}} & 0.62421 \\
    \end{tabular} &
    \begin{tabular}{cc}
        \textbf{{German\_XLM}} &  0.61658 \\ 
        \textbf{{German\_BERT}} & 0.63514 \\
    \end{tabular} \\ \hline
    
     \textbf{French (FR)} & 
        \begin{tabular}{cc}
        
       \textbf{{French\_XLM}} &  0.61209 \\
       \textbf{{French\_BERT}} & 0.58065 \\
    \end{tabular} &
    \begin{tabular}{cc}
        \textbf{{French\_XLM}} &  0.56994 \\
        \textbf{{French\_BERT}} & 0.60932 \\
    \end{tabular} \\ \hline
    
     \textbf{Chinese (ZH)} & 
        \begin{tabular}{cc}
       \textbf{{Chinese\_XLM}} &  0.62267 \\
       \textbf{{Chinese\_BERT}} & 0.54087 \\
    \end{tabular} &
    \begin{tabular}{cc}
        \textbf{{Chinese\_XLM }} &  0.54025 \\    
        \textbf{{Chinese\_BERT}} & 0.61938 \\
    \end{tabular} \\ \hline
    
     \textbf{Japanese(JA)} & 
        \begin{tabular}{cc}
       \textbf{{Japanese\_XLM}} &  0.60377 \\
       \textbf{{Japanese\_BERT}} & 0.51262 \\
    \end{tabular} &
    \begin{tabular}{cc}
        \textbf{{Japanese\_XLM}} &  0.52043 \\     
        \textbf{{Japanese\_BERT}} & 0.59127\\
    \end{tabular} \\ \hline

        \hline
    \end{tabular}
    \end{adjustbox}
    \caption{Star Rating Prediction}
    \label{tab:SRPt}
\label{table:SRP}
\end{table}

It can be observed that machine translation didn’t affect if not significantly improve or derail performance of models in their downstream applications for languages like Spanish, German and French, which as clearly European languages and share a lot of semantic similarities with english. On the other hand Machine Translation significantly, affects the performance of languages Chinese and Japanese in a negative way, as they are significantly unique to english. However no model could reach the fine tuned performance of the english baseline of about 0.91 for english reviews, which could be due to gaps in Machine Translation models, as theoretically they translated sentences in english should be able to convey full meaning of original sentence and reach english benchmarks.However failure to do so even after intensive fine-tuning, leads us to conclude that either more robust fine-tuning on a much larger dataset rather than a subset or a more robust pipeline should help, which would have been outside the scope of our project given the resource, and time.

\section{Observations}
Table 1 and Table 2 show the F1 scores for our tasks for each of the language specific model before and after machine Translation.

We can infer that XLM-RoBERTA performs slightly better than BERT for each language dataset as expected due to XLM’s cross-lingual capabilities. For the task of Sentiment Analysis, the average F1 score over all the models was 0.89. Before Machine Translation, the German model got the best F1 score across all the models, while the Chinese performed the worst. After Translation, the performance of all the models degraded with the Japanese Model being the most affected. The Spanish model got the best F1 score after machine Translation.
For the task of Star Rating Prediction, the average F1 score over all the models was 0.61. Before Machine Translation, the German model outperformed all the models, while the Japanese performed the worst. After Translation, the performance of all the models degraded with the Japanese and Chinese Models being the most affected. The German model still got the best F1 score after machine Translation across all the models.

\section{Conclusion and Future Work}

During the course of the project, we developed language-specific models and models utilizing translated texts. There was no significant difference between the models developed during the project(language-specific and models using translated tests) as they achieved similar performance. However, the slight difference that occurred can be due to the following shortcomings. 

Languages like Spanish, German, and French share many similarities with English regarding sentence structure sharing the same SVO word order with English. As a result, machine translation from these languages to English may preserve the original meaning well, leading to consistent sentiment analysis results. However, Asian languages like Japanese and Chinese have different linguistic structures. Japanese syntax follows SOV word order while Chinese sentence structure is characterized by its lack of inflectional morphology and grammatical markers, relying heavily on word order and context for conveying meaning. Machine translation may struggle to accurately capture such semantic meaning, leading to loss of information and thus Higher \textbf{Semantic difference}.

Asian cultures, for example, may have unique ways of conveying sentiment that differ from Western cultures.So \textbf{Cultural difference} also plays a role in Machine Translation.
Lastly, The \textbf{availability and quality of training data} may vary across languages. English sentiment analysis models may have been trained on larger and more diverse datasets compared to models for other languages. This discrepancy in training data quality can impact the effectiveness of sentiment analysis after machine translation, especially for languages with less available data.

In summary, the effectiveness of machine translation in preserving sentiment and maintaining performance in downstream applications such as sentiment analysis depends on factors such as linguistic similarity, syntactic complexity, cultural differences, and data availability. While machine translation may perform well for languages closely related to English, it may encounter challenges in accurately capturing sentiment for languages with greater linguistic and cultural differences.

To improve the work done, there can be further experimentation, which involves, firstly, fine-tuning machine translation models specifically for sentiment-related tasks. This could involve adding sentiment-specific data or annotations into the fine-tuning process to improve the efficacy of translations, especially for languages with high linguistic differences from English. Extensive literature is available to improve sentiment analysis of models by training them on general or domain-specific Knowledge graphs\cite{Li2023-rg}, such as ConceptNet\cite{Speer2016-hg}. Secondly, utilizing multimodal approaches incorporating visual and textual information for sentiment analysis across languages. Exploring how images or videos can complement machine-translated text to improve the performance of sentiment analysis, especially in languages where textual data may be limited or unreliable. Some preliminary work on this could be explored in work done by Yoon et al. \cite{Yoon2018-tm}. Finally, focusing on improving sentiment analysis performance in low-resource languages with a lack of training data. Experimentation can be done by exploring transfer, semi-supervised, or unsupervised learning to adapt to sentiment analysis tasks for languages with limited labelled data.

In summary, the effectiveness of machine translation in preserving sentiment and maintaining performance in downstream applications such as sentiment analysis depends on factors such as linguistic similarity, syntactic complexity, cultural differences, data availability, and translation quality. While machine translation may perform well for languages closely related to English, it may encounter challenges in accurately capturing sentiment for languages with greater linguistic and cultural differences.
\section{Code Availability}

We have published all our code files and processed the dataset to 
\href{https://github.com/Nalin21478/NLP-Project}{GitHub}. Fine-tuned model checkpoints of all the models can be found at \href{https://drive.google.com/drive/folders/1R00drvoxtaUIxxDaJxbdy2yfBKd2mkQ5?usp=sharing}{Drive}.

{\small
\bibliographystyle{ieee_fullname}
\bibliography{bibl}
}

\end{document}